\documentclass[article, twocolumn, notitlepage,superscriptaddress, longbibliography]{revtex4-2}
\usepackage{natbib}
\usepackage{graphicx}
\usepackage{amssymb}
\usepackage{amsmath}
\usepackage{ifthen}
\usepackage{braket}
\usepackage{xcolor}
\usepackage{bm}
\usepackage{bbm}
\usepackage{bbm}
\usepackage{maybemath}
\usepackage{comment}
\usepackage{siunitx}
\usepackage{float}
\usepackage{tabularray}
\usepackage{dcolumn}
\newcolumntype{d}[1]{D{.}{.}{#1}}
\usepackage{subfigure}

\definecolor{garrosgreen}{rgb}{0.1, 0.4, 0.1}
\definecolor{dartmouthgreen}{rgb}{0.05, 0.5, 0.06}
\definecolor{duelferred}{rgb}{0.7, 0.2, 0.1}
\definecolor{cambridgeblue}{rgb}{0.1, 0.3, 1.0}
\definecolor{oxfordblue}{rgb}{0.05, 0.2, 0.7}

\newcommand{\addrA}{Department of Chemistry, Physics and Materials Science, Fayetteville State University, Fayetteville, NC 28301, USA}
\newcommand{\addrB} {Department of Mathematical and Statistical Sciences, University of Nebraska Omaha, Omaha, NE 68182, USA}
\newcommand{\addrC}{Jack Britt High School, Fayetteville, NC 28306, USA}

\begin{document}

\title{Machine learning and machine learned prediction in chest X-ray images}

\author{Shereiff Garrett} % M. refers to middle initial
%\thanks{These authors contributed equally.}
\affiliation{\addrA}

\author{Abhinav  Adhikari }
%\thanks{These authors contributed equally.}
\affiliation{\addrB}

\author{Sarina Gautam}
\affiliation{\addrC}

\author{Da'Shawn M. Morris}
\affiliation{\addrA}

\author{Chandra M. Adhikari}
\email[ Corresponding author:\;]{cadhikari@uncfsu.edu}
\affiliation{\addrA}

\begin{abstract}
Machine learning and artificial intelligence are fast-growing fields of research in which data is used to train algorithms, learn patterns, and make predictions. This approach helps to solve seemingly intricate problems with significant accuracy without explicit programming by recognizing complex relationships in data. Taking an example of  5824 chest X-ray images, we implement two machine learning algorithms, namely, a baseline convolutional neural network (CNN) and a DenseNet-121, and present our analysis in making machine-learned predictions in predicting patients with ailments. Both baseline CNN and DenseNet-121 perform very well in the binary classification problem presented in this work. Gradient-weighted class activation mapping shows that DenseNet-121 correctly focuses on essential parts of the input chest X-ray images in its decision-making more than the baseline CNN. 
\end{abstract}

\maketitle

%\noindent{\textbf{Keywords}:  Machine learning; Artificial intelligence; Convolutional neural network; DenseNet121; Pneumonia detection; Grad-CAM} %\\ 
%\tableofcontents

\section{\label{sec:level1}   INTRODUCTION}

Systems that can be expressed and explained by known physical, mathematical, or logical laws are better understood through theoretical models, as they offer precise, interpretable, and generalizable predictions, providing direct insight into cause-and-effect relationships. Some systems are so intricate, with unclear dynamics of their parameters, that presenting input-output relationships in a closed form of a mathematical model is difficult or not feasible at all, or theories are impractical and/or incomplete. As machine learning (ML) can adapt to real-world messiness and make predictions using Artificial intelligence (AI), a data-driven approach can be a wise choice to tackle such a system, provided a sufficiently large dataset is available. Disease diagnosis is one of many areas where ML and AI have great potential to revolutionize the diagnosis process by reviewing immense amounts of images and performing image classification.

Neural networks (NNs), inspired by the human brain, have demonstrated human-level performance across multiple task domains, although the NN is based on statistical measures and relies on human-simulated intelligence programmed and controlled through algorithms. One node of an NN mimics the brain's smallest measured unit, such as a voxel. The input layer of the NN mimics the raw data received by sensory organs and/or the primary cortex, such as the eyes, skin, or ears. NN's hidden layers imitate the intermediate processing of the brain, as done by the neocortex, to extract hierarchical features analogous to the deeper processing steps that occur in the brain. Output layer of NN copies the functionality of motor cortex, aka brain's decision regions, to produce final action or classification  akin to behavior/output in brain~\cite{NaMaAoKa2021}

The article aims at two objectives. First, we briefly review two different approaches to NNs: the baseline Convolutional Neural Network (CNN)~\cite{MaLeMi2019} and a densely connected DenseNet-121 CNN~\cite{GaZh_IEEE_2018}, used in this study, which takes advantage of an open-source deep-learning framework called PyTorch~\cite{Paszke_2019_PyTorch}, a Python library. PyTorch utilizes tensors as a fundamental data structure, enabling researchers to modify the network’s behavior in real-time. Second, we implement these neural network techniques to analyze chest X-ray images with the aim of accurately predicting the presence of an ailment in the chest and demonstrating the capabilities of ML and AI.

Pneumonia is one of the leading causes of illness and death worldwide, especially in more vulnerable populations~\cite{BoGrMaLu2025, SaScAs2021, ObMuAd2025}. According to the "National Center for Health Statistics" report of the Centers for Disease Control and Prevention (CDC) for 2023,  pneumonia is the eleventh leading cause of death in the USA. For the 0-19 age group, it was listed as the 9th leading cause of death and the second among the disease-related deaths behind COVID-19 in the year 2019 ~\cite{FlWhSe2023}. In 2023, approximately 1.5 million patients visited emergency departments with pneumonia as the primary diagnosis, caused by infectious organisms, and pneumonia resulted in the deaths of 41,210 patients, which is 1.23 per 10,000 population~\cite{CDC_2023_Pneumonia}.

%Racial distribution on pneumonia-related mortality shows that non-hispanic black americans has the highest pneumonia-related mortality. For example the Age-adjusted mortality rates (AAMR) ratio for non-hispanic black americans is  approximately 1.11 times greater than white americans in 2016-2018~\cite{BoGrMaLu2025}. Though AAMRs have declined since 1999 (from ~22.9 to ~9.9 per 100,000), the relative gap between Black and White populations has remained particularly affecting Black Americans~\cite{ObMuAd2025}
%National Center for Health Statistics (NCHS) data brief report for 2016-2018 shows that non-Hispanic Black Americans face both significantly higher mortality and greater hospitalization/ED usage due to pneumonia compared to other groups~\cite{SaScAs2021}. 

A Rapid and accurate diagnosis from chest X-rays is critical and can be challenging. It is worth noting that deep Learning has shown considerable promise in automating and supporting the clinical interpretation of X-ray images. Reliable and interpretable AI tools can assist radiologists in accurately identifying any ailment present in images, aiding in diagnosis.

The baseline CNN is a custom-designed neural network that utilizes convolutional layers for feature extraction and fully connected layers for classification. It uses a gray image of dimensions  $ 1\times 32\times 32$ or a color image of dimensions $ 3\times 32\times 32$, where the first numbers 1 and  3 denote the number of channels (1 for gray and 3 for RGB), while 32s are the height and width of the images in pixels. A densely connected CNN, such as DenseNet121, however, has an input size of $3\times 224\times 224$. It features a deep and pre-trained architecture with multiple layers. In this project, we compare the effectiveness of these two deep learning models for pneumonia detection on chest X-ray images, assessing both classification performance and model interpretability.

%%%%
\section{\label{sec:level2}   Methods} 

Taking the chest X-ray dataset from Kaggle, which contains 5824 JPEG images with two labels, namely, NORMAL and PNEUMONIA~\cite{chest-x-rays-qjmia}, we investigate the effectiveness of ML models to predict pneumonia images correctly. This work will support researchers focusing on automating ML/AI-based methods to detect and classify human diseases from medical images.

All low-quality or unreadable scans of chest X-rays were removed to have a better data-driven analysis. The dataset was split into training, validation, and test sets for fair evaluation, with distributions of 88\%, 8\%, and 4\%, respectively. Preprocessing involves the application of auto-orientation, resizing, and normalizing each image to the same size. Images were augmented, allowing the saturation, brightness, and exposure to vary by only 5\% to above or below the values of the original X-rays. These image data were used to train separate baseline CNN and DenseNet 121 deep learning architectures and were evaluated using various metrics, including confusion matrices~\cite{SaRo2024, St1997} , receiver operating characteristic (ROC) curves~\cite{JuDe2018}, and the area under the curve (AUC) in ROC curves~\cite{Na2022}.  

\begin{figure}[tbh!]
\includegraphics[width=0.495\textwidth]{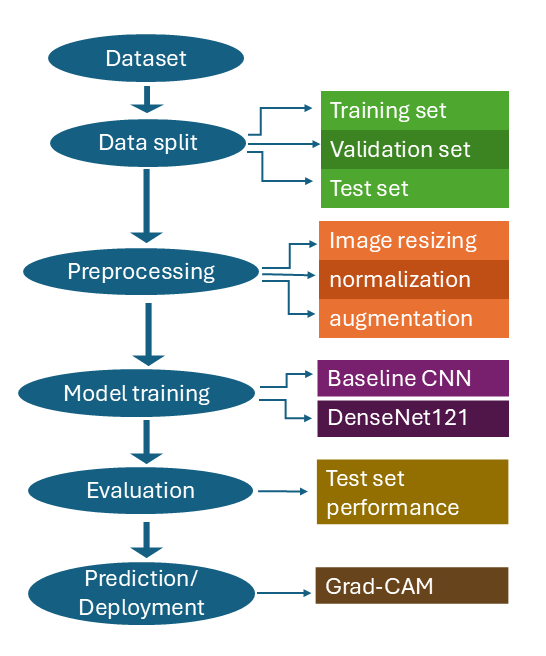}
\caption{\label{fig1} Work flow. }
\end{figure}

The test performance was tested for both models. To interpret the observations and predictions done by these models, the Gradient-weighted Class Activation Mapping (Grad-CAM)~\cite{SeCo2020} model interpretability technique was used. Grad-CAM exploits the gradients of the target class percolating through the final convolutional layer to create a heat-map highlighting important regions in the input image. Figure~\ref{fig1} shows the workflow followed in this binary classification with ML and AI. Before presenting the results, a brief discussion on each of the NNs employed to learn chest X-rays in this article is in order.

\subsection{\label{sec:level2a}   Baseline CNN} 

A baseline CNN is a simple yet effective neural network model used to establish a minimum standard for performance in a specific deep learning task, typically in image classification, object detection, segmentation, and/or segmentation. It usually acts as a benchmark for evaluating more complex models, not only by providing insight into whether deeper models are really helping, but also helps to validate data pipeline, loss functions, and training setup, and in turn provides a starting point for improvement of model via architecture tuning, augmentation, regularization, and hyperparamenter adjustment if needed.

The baseline CNN follows the sequential connectivity in such a way that each layer builds upon the immediately preceding feature map. Consequently, the input in the $n$th layer can be written as
\begin{align}
x_n= H_n \left( x_{n-1}\right)\,,
\end{align}
where $H_n$ is a composite function applied at $n$th layer. The basic building block of baseline CNN's
$H_n$ function is Conv $\rightarrow$ ReLU $\rightarrow$ Pooling, where Conv and ReLU denote convolution and rectified linear unit, respectively. The Conv extracts spatial patterns such as edges, textures, etc, ReLU enables the model to learn non-linear representations, and Pooling reduces spatial dimensions and overfitting.

\subsection{\label{sec:level2b}  DenseNet-121 }

Unlike baseline CNNs, DenseNet connects each layer to every other layer in a feed-forward fashion such that input $x_n$ to each layer $n$ receives feature maps from all preceding layers~\cite{GaZh_IEEE_2018}. The architecture of each layer, receiving input from all previous layers, promotes feature reuse and efficient gradient flow. DenseNet connectivity formula can be written as
\begin{align}
x_n= H_n \left(\left[ x_0, x_1, x_2, \cdots, x_{n-1}\right]\right)\,,
\end{align}
where $H_n$ is a composite function consisting of batch normalization (BN), ReLU, and convolution. DenseNet features a bottleneck design (BN $\rightarrow$ ReLU $\rightarrow$ Conv $1\times 1$) preceding the main $3\times 3$ convolution, aiming to reduce the number of input channels and thereby decrease computational cost while increasing the depth of the feature map without a significant increase in parameters. 

The DenseNet-121 architecture consists of 121 layers, comprising an initial convolution, four dense blocks, three transition layers, and a fully connected (FC) output layer. The four dense blocks have 6, 12, 24, and 16 pairs of bottleneck layers with 256, 512, 1024, and 1024 output channels, respectively. %Each dense block is a composite of two convolutional layers. 
The  layers count as follows: 1 initial convolution + 2$\times$ ( 6 dense block-1 + 12 dense block-2 + 24 dense block-3 + 16 dense block-4) + 3 transitions + 1 FC to make 121.

One may wonder whether the baseline CNN can outperform DenseNet-121. A large number of layers makes networks of baseline CNN deeper, which can model complex functions, increasing the receptive field and abstraction level. Thus, theoretically, matching or even exceeding the DenseNet-121 in accuracy from a baseline CNN with a large number of layers is possible. However, a simple deep CNN usually fails due to various reasons. First, gradients are back-propagated layer by layer in deep NN, as a result, the gradients can vanish in early layers, resulting in ineffective learning, or the gradient blows up, making unstable updates. Increasing the depth makes the network weaker at propagating the learning signals back to early layers. Second, each layer of the baseline CNN discards the features of the previous layers, causing it to learn again from scratch. Thus, making the baseline CNN deeper is essentially relearning a similar pattern repeatedly, which wastes learning capacity and makes training even more challenging. Third, increasing the depth in baseline CNN increases the number of convolutional filters, each with its own weights, causing a rapid growth of the memory usage, leading to a slow training process. In summary, simply increasing the layers in the baseline CNN does not make it smarter than DenseNet-121, as increasing the layers introduces many architectural issues that deteriorate its performance. Instead, using dense connections, the DenseNet-121 design performs smartly, maximizing gradient flow, feature reuse, and parameter efficiency.

\begin{figure*}[tbh!]
\includegraphics[width=0.95\textwidth]{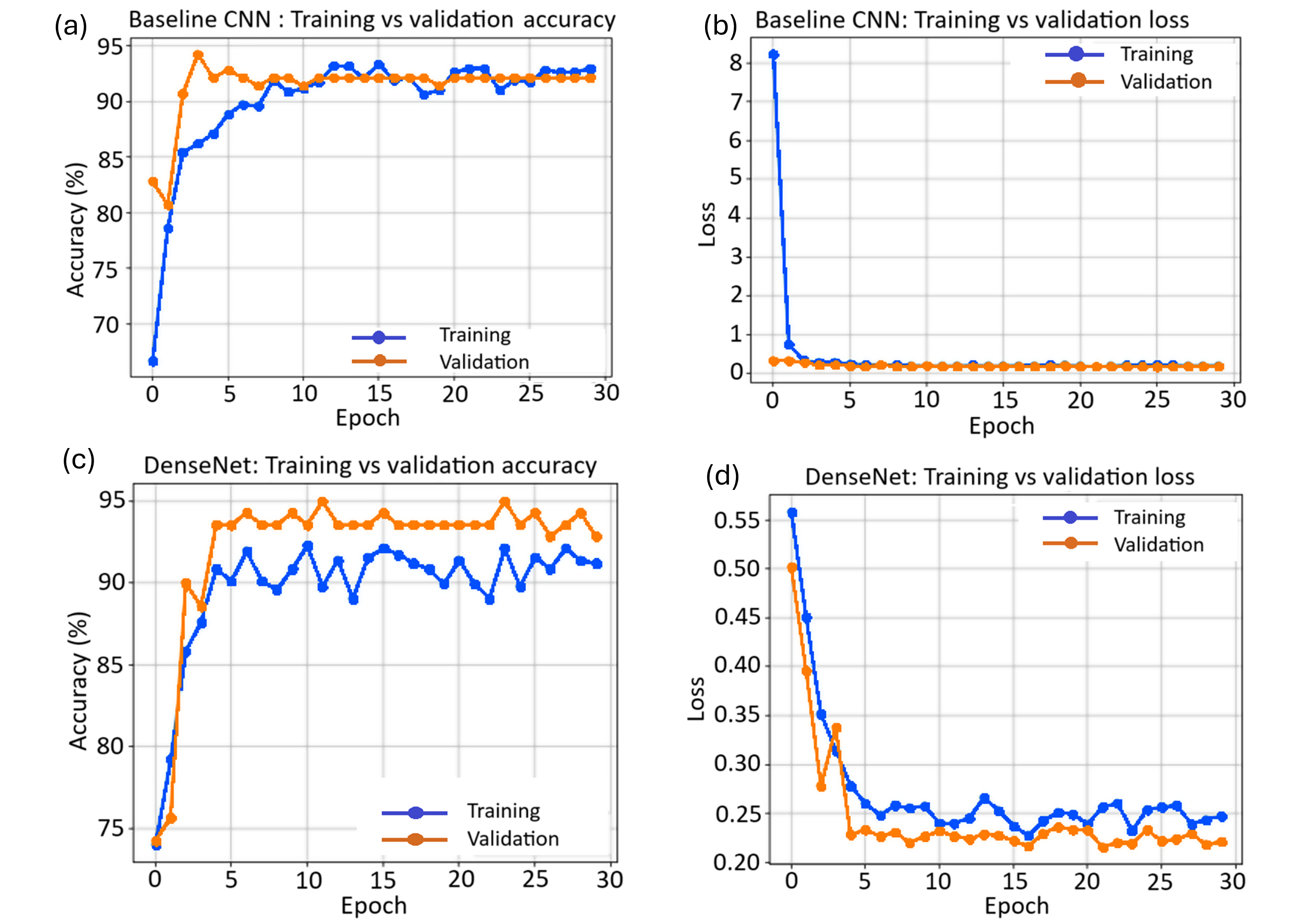}
\caption{\label{figTVEpochs}  Training and validation accuracies and losses as a function of epochs for baseline CNN and Densenet-121.   }
\end{figure*}

\section{\label{sec:level4}   Non-linearity: A Key in Neural Network}

Although linear models are easy to train and interpret due to the proportional or hyperplane relationships between input and output, the patterns that field-collected data in real-world systems often exhibit are complex in nature and non-separable by linear boundaries, making non-linearity a key factor to consider in their study. Without any activation and incorporation of non-linearity, an NN  is essentially a single-layer linear model, regardless of the NN's depth~\cite{FMWi2024}. Consequently, a non-linear model that is absent or suppressed faces severe limitations on its ability to solve real-world challenges, such as image classification, medical diagnosis, and facial detection. Non-linearity enables NNs to evolve into universal function approximators, empowering them to model complex patterns and dependencies that exist among many features, the non-linear decision boundaries the system has, and high-dimensional data relationships that arise from the interactions among many input features.

In NNs, non-linearity is introduced through activation functions such as sigmoid, Tanh, ReLU, Swish, Mish, softplus, exponential linear unit (ELU), etc, which provoke bending and curvature in the NN's computational process. For example, the sigmoid function $f(x) = (1+e^{-x})^{-1}$ provides $S$-shaped saturation varying value in the range of 0 to 1. Tanh function $f(x) =\tanh(x)= (e^{x} - e^{-x})/(e^x + e^{-x})$ gives a similar-shaped activation to that of sigmoid but ranges from -1 to 1, centering at zero. ReLU function $f(x) = \text{max}(0, x)$ introduces sparse and non-saturating non-negative activation in the range of 0 to $\infty$. Swish $f(x)= x/ (1+e^{-x})$ and Mish $f(x)= x\cdot \tanh\left[\ln(1+e^x)\right]$ activation functions give smooth activation in the range of $-\infty$ to $\infty$. Softplus is a smooth approximation of ReLU. In need of a zero-centered smooth non-saturating function, ELU defined by  $f(x)=x, \;\;\text{if}\;\; x\ge 0$ and $f(x) =\alpha (e^{x}-1)$ otherwise, is used. For reference, see~\cite{DuSiCh2022} for details on activation functions.

Non-linear activation functions embedded in the composite function $H_n$ transform the weighted sum of inputs in a neuron into an output that is no longer linear in relationship. Non-linearity empowers the NN to solve real-world, linearly non-separable, complex patterns, thereby expanding its representational capability. Furthermore, non-linearity enables NN hierarchical feature learning, extracting abstract features from all previous layers in DenseNet-121 or just the previous one in baseline CNN.

\section{\label{sec:level5}   Results and discussion }

Data were analyzed using an open-source ML library called PyTorch, developed by Meta AI, used for building and training deep learning models designed from first principles to support an imperative and Python programming style~\cite{Paszke_2019_PyTorch}. To diagnose how the model's performance evolves over training epochs, we visualized the comparison of training and validation accuracy and loss versus epochs, as presented in Fig.~\ref{figTVEpochs}. Accuracy measures the ratio of total correct predictions to the total predictions made by the model. It can be expressed as a percentage by multiplying it by 100. The validation accuracy is slightly improved, resulting in a reduced loss with the DenseNet-121 model compared to the baseline CNN. The loss is a measure of error. It assesses how bad the models' predictions are. Note that accuracy and loss need not sum to unity or 100\%. The comparative analysis of Figs.~\ref{figTVEpochs}(b) and \ref{figTVEpochs}(d) shows that DenseNet-121 is marginally better than the baseline CNN.

\begin{figure*}[tbh!]
\includegraphics[width=0.975\textwidth]{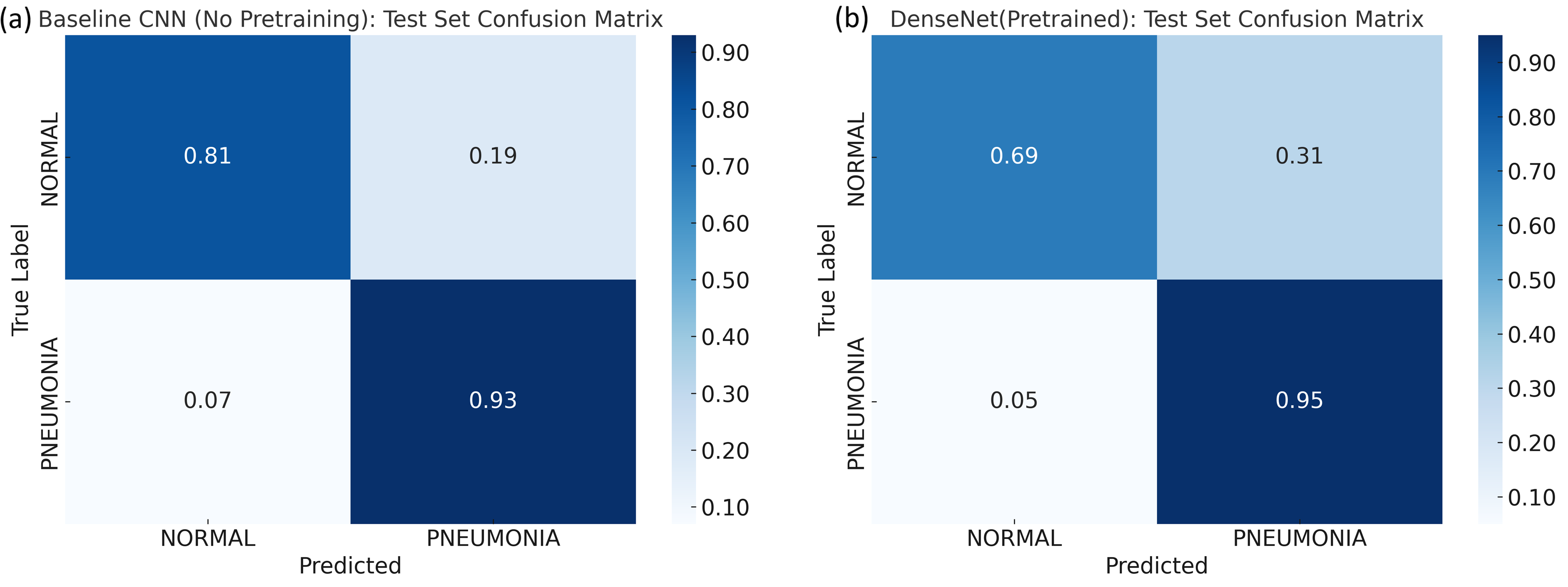}
\caption{\label{TpFn} Confusion matrices, for (a) baseline CNN and (b) DenseNet-121, showing the values of predicted label for each true label normalizing the values for each class to unity. }
\end{figure*}

We have used confusion matrices, also known as error matrices, as a performance evaluation tool for our binary classification problems. The diagonal entries of the matrices (true positive and false negative) represent all instances in which the model correctly predicted the classes, namely, "NORMAL" and "PNEUMONIA". The off-diagonal entries (false positives and false negatives) indicate the confusion level or error in prediction, where one class is mislabeled as another. The baseline CNN correctly predicted 81\% of the normal chest X-rays as normal and 93\% of the pneumonia cases as pneumonia. 19\% of normal cases were falsely predicted as pneumonia, and 7\% of the pneumonia cases were also predicted as normal (see Fig.~\ref{TpFn}). Predicting normal chest X-rays as normal is 12\% less for DenseNet-121 than baseline CNN, while the DenseNet-121 predicted pneumonia cases as pneumonia slightly better than baseline CNN.

\begin{figure*}[tbh!]
\includegraphics[width=0.975\textwidth]{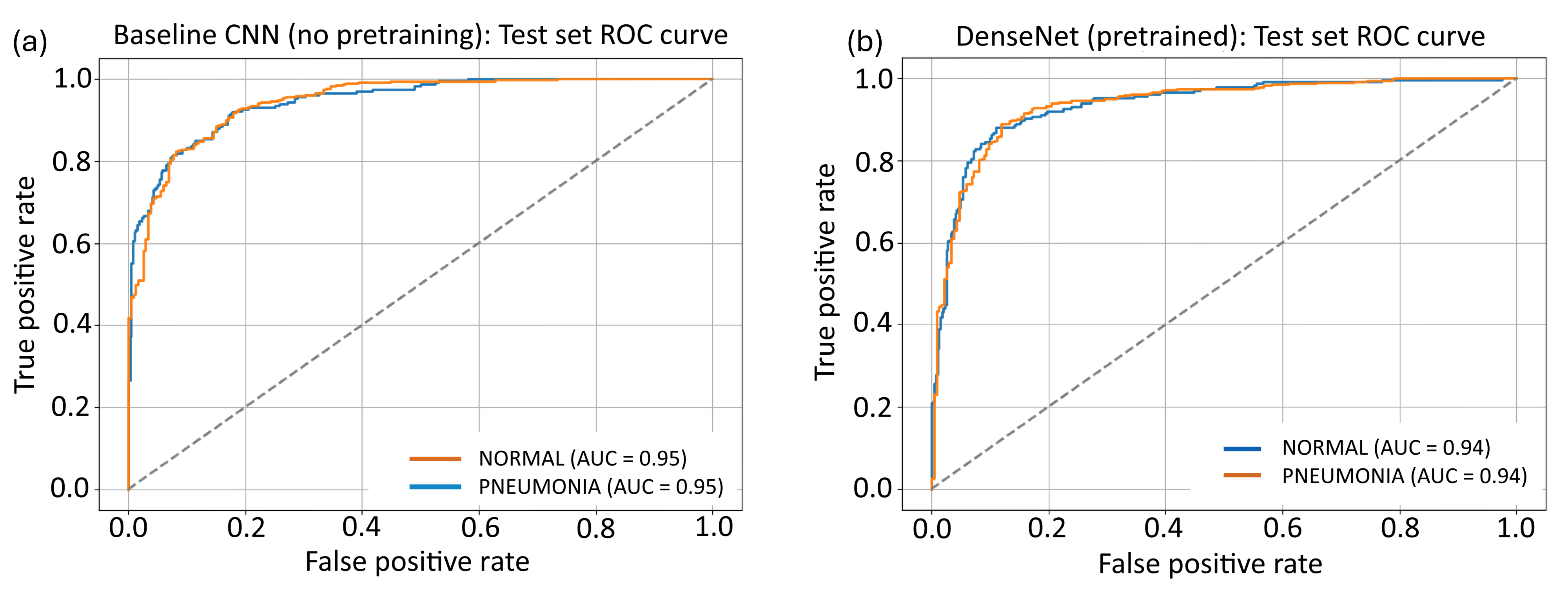}
\caption{\label{testsetPerformance} Receiver operating characteristic (ROC) curves showing true positive rate (TPR) on the vertical axis and false positive rate (FPR) on the horizontal axis for baseline CNN (a) and DenseNet-121 (b), respectively.  Numbers in the legend are area under the curve (AUC). The dashed line is for random guessing with AUC = 0.5.}
\end{figure*}

The ROC curves render a visual and quantitative measure to examine the ability of a model to differentiate between classes across all possible classification thresholds. The ROC curves visualize the variation of true positive rate (TPR) with false positive rate (FPR), where TPR =  TP/ (TP+FN) and FPR =  FP/ (FP+TN) with TP, FP, TN, and FN being true positive, false positive, true negative, and false negative counts, respectively~\cite{OmSoAl2023, BuBr1988}. TP count measures how many of the predicted positives are truly positive, while FP count measures how many of the predicted positives are actually negative. The area under the curve (AUC) abridges the model's overall ability to distinguish its constituent classes into a single number. Any model with an AUC of less than 0.5 is considered a bad model, while a model with an AUC of 1.0 is a perfect model that flawlessly discriminates its classes. In real-world practice, an AUC of 1.0 is only theoretically feasible, and models with an AUC of 1.0 may indicate issues such as data leakage or overfitting, or an algorithmic error. It is also important to note that ML model struggles or overfit when data is scarce, and overfitting is always a red flag in ML. Any models with an AUC greater than 0.9 are considered excellent-performing models, with higher values indicating better performance. Figure~\ref{testsetPerformance} shows the ROC curves for baseline CNN and DenseNet-121 models. The AUC values for both cases fall within the excellent performing range, indicating that both models are working excellently in distinguishing between chest X-rays of normal patients and those with pneumonia.


\begin{figure*}[tbh!]
\includegraphics[width=0.95\textwidth]{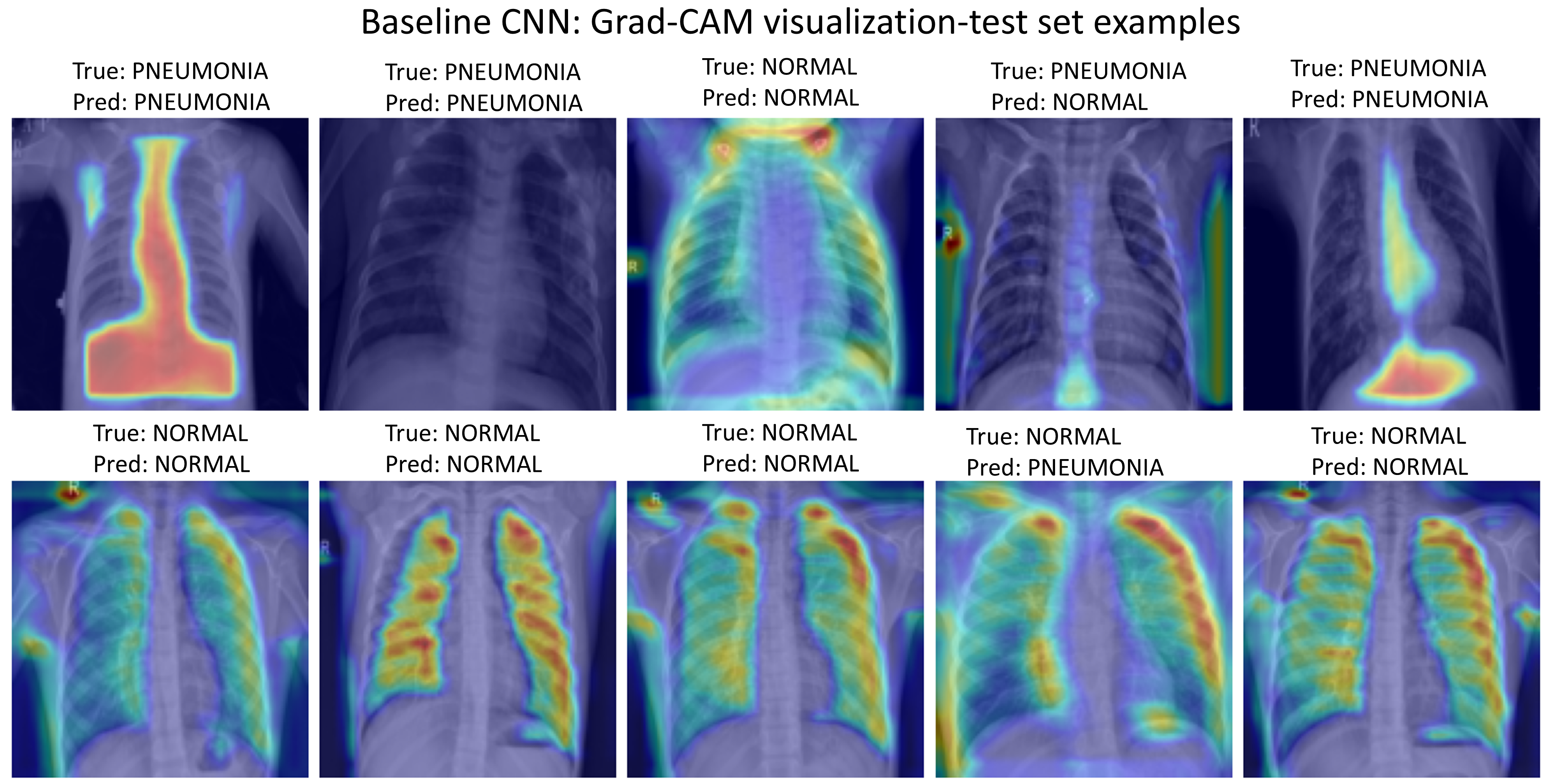}
\caption{\label{cnn-gradcam} Some example images of Grad-CAM visualization in baseline CNN. Bright red spots are strongly influential and  dark-blue are less or no influential areas. }
\end{figure*}

\begin{figure*}[tbh!]
\includegraphics[width=0.95\textwidth]{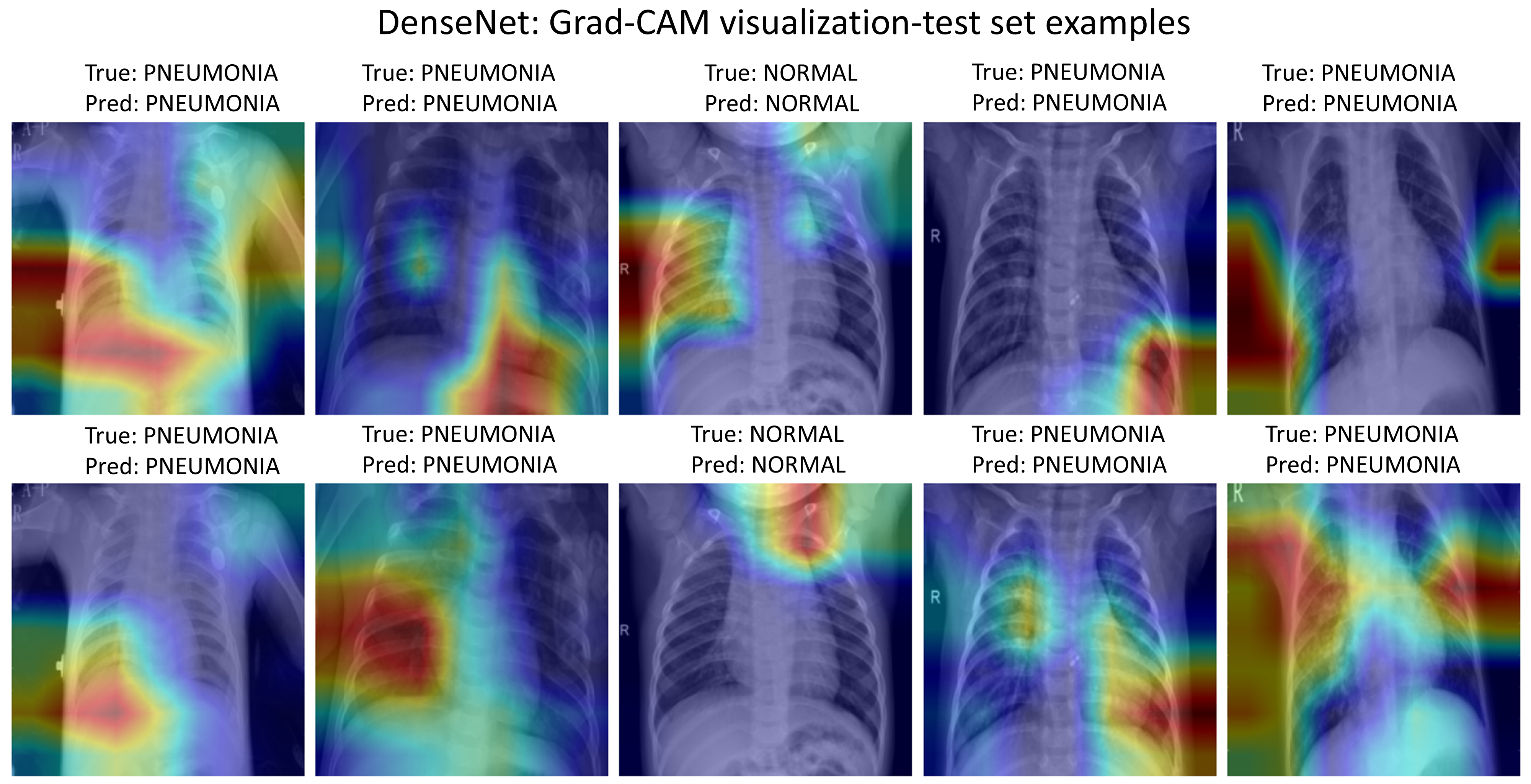}
\caption{\label{densenet-gradcam} Some example images of Grad-CAM visualization in DenseNet-121. Bright red spots are strongly influential and  dark-blue are less or no influential areas. }
\end{figure*}

\begin{figure*}[tbh!]
\includegraphics[width=0.95\textwidth]{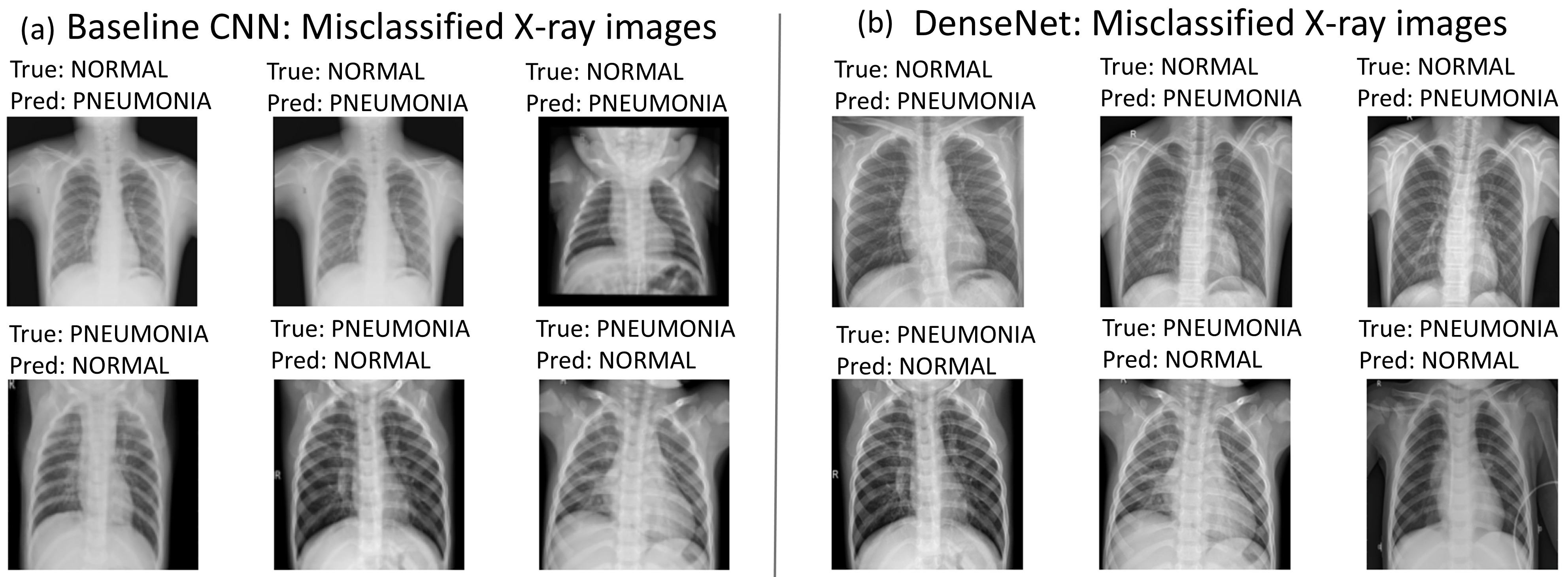}
\caption{\label{missclassified} Some randomly selected misclassified chest X-ray images from (a) baseline CNN and (b) DenseNet-121 models. }
\end{figure*}

A NN learns features automatically from raw data, adjusting its internal parameters to map inputs to correct outputs. Grad-CAM is a powerful visualization tool that generates a heatmap highlighting the critical regions in an image, aiding in class prediction. The Grad-CAM is based on gradients of the class score with respect to feature maps. As the input image goes through the NN, the model collects feature maps from a convolutional layer and evaluates the gradient of the output score of the target with respect to the selected feature maps. The NN computes weights and averages out gradients spatially to get important weights for each feature map. Then, weighted activation maps are created, and an overlay heatmap is applied to the original image. Strongly influential areas are represented by bright red, while less influential or non-influential regions are represented by dark blue. Grad-CAM helps to understand why the model predicted a particular class, which not only reveals which parts of the input image were most important for a class decision, but also builds trust in AI decisions by providing the reason behind the prediction. Both baseline CNN and DenseNet-121 can localize regions of interest in chest X-rays, while DenseNet-121 produces more focused and clinically plausible attention maps (see Figs.~\ref{cnn-gradcam} and \ref{densenet-gradcam}). In baseline CNN, being trained from scratch, Grad-CAMs are more diffuse and sometimes highlight irrelevant areas, indicating less precise feature learning and, in some instances, misfire even on true positives. In contrast, the DenseNet-121 demonstrates higher consistency in correct predictions and focuses on meaningful lung zones, highlighting disease-relevant regions more clearly, supporting better interpretability for pneumonia detection.

NN is a powerful means of learning from data and making decisions to classify classes present in the dataset correctly. However, NN models can still misclassify due to various reasons such as visual and statistical similarity of data classes, imbalanced data, underfitting or overfitting, wrong hyperparameters, sensitivity of the data to noise, class overlap, insufficient data, noisy training data, and so on~\cite{NaZhFi2023}. Some randomly selected Baseline CNN and DenseNet-121 misclassified chest X-rays are shown in Fig.~\ref{missclassified}. The chest X-ray dataset is large enough, with more than 5000 images for only two classes. Data has been split into the well-accepted data set distribution in NN. There is no class overlap, and we did not perform hyperparameter tuning in the model. The misclassification in our model is mainly due to the inherent visual similarity in the images, despite their actual dissimilarity, as they depict different people taken at different times by different professionals.

It is essential to note that, although deep learning NN models, as presented in this work, such as the baseline CNN and DenseNet-121, can be astonishingly accurate, human experts are still crucial for reviewing, validating, and contextualizing AI results before accepting the predictions made by them, as AI lacks real-world understanding and clinical judgment and cannot make judgments about whether inputs are out-of-distribution. For a given input, models will still provide a prediction. Still, the conditions of the data may have changed, such as the equipment being different or the patient suffering from a different but related disease. Furthermore, the model may make a correct prediction with high confidence, but if it looks at the text or a wrong spot, rather than the organ to focus on,  in this case, the lungs. A field expert can recognize exceptions, identify new patterns, and detect critical errors but algorithms may fail on rare or unexpected inputs.

\section{  Conclusion}

The architecture of two NN models, namely, baseline CNN consisting of only a few layers and DenseNet-121 comprising 121 layers, is briefly reviewed. A discussion on the power of nonlinearity in NN is presented. Using a publicly available dataset of chest X-rays containing over 5,000 images, the capabilities of baseline CNN and DenseNet-121 NNs were analyzed in detecting images of patients with ailments.

Both the baseline CNN and DenseNet-121 models accurately detected patients with pneumonia. To reach this conclusion, we observed variations in training and validation accuracies and losses as a function of epoch, confusion matrices, and ROC curves. We used  Grad-CAMs as a tool to visually explain the predictions of  NNs and better understand why the model predicted a certain class for a chosen image.   In baseline-CNN, Grad-CAMs are more dispersed and sometimes highlight irrelevant areas, indicating less precise feature learning and, in some instances, misfire even on true positives. The DenseNet-121 demonstrates higher consistency in correct predictions and focuses on meaningful lung zones, highlighting disease-relevant regions more clearly, which supports better interpretability for pneumonia detection, making it more logical from a human perspective.

AI models make predictions even if some inputs may fall into out-of-distribution. Being a data-driven model, ML and ML-based decisions are based on data that trains the model, rather than the domain knowledge that a human expert adds. NN models serve as powerful tools, assisting in prediction and informed judgment, and human experts make certain that their use is safe, responsible, and intelligent.

%\begin{acknowledgments}
%This work is supported by the Department of XXXX Grant number XXXXXXX. 
%\end{acknowledgments}

\typeout{} 
%\bibliographystyle{aipnum}
%\bibliography{myBibFile}
%

%
\end{document}